\documentclass[letterpaper, 10 pt, conference]{ieeeconf}  

\IEEEoverridecommandlockouts                              

\usepackage{amsmath} 
\usepackage{graphicx}
\usepackage{longtable}
\usepackage{supertabular}
\usepackage{hyperref}
\usepackage{booktabs}
\usepackage{amssymb}  
\usepackage{pdflscape}
\usepackage{tablefootnote}
\usepackage{threeparttable} 
\usepackage{arydshln} 

\title{\LARGE \bf
Datasets on object manipulation and interaction: a survey
}

\author{Yongqiang Huang and Yu Sun
\thanks{The authors are with the Department of Computer Science and Engineering at the University of South Florida, Tampa, FL, USA. email: \texttt{yongqiang@mail.usf.edu, yusun@cse.usf.edu}.}%
}

\newcommand{\sd}{\cite{pham2009}}
\newcommand{\mmac}{\cite{torre2009}}
\newcommand{\gaze}{\cite{fathi2012}} 
\newcommand{\gazep}{\cite{fathi2012}+} 
\newcommand{\mpiic}{\cite{rohrbach2012}}
\newcommand{\mpiip}{\cite{rohrbach2012eccv}}
\newcommand{\mpiid}{\cite{rohrbach2015}}
\newcommand{\salad}{\cite{stein2013}}
\newcommand{\ace}{\cite{shimada2013}}
\newcommand{\yc}{\cite{das2013}}
\newcommand{\kthc}{\cite{pieropan2014}}
\newcommand{\bb}{\cite{kuehne2014}}
\newcommand{\tumk}{\cite{tenorth2009}}
\newcommand{\radl}{\cite{messing2009}}
\newcommand{\opp}{\cite{roggen2010}}
\newcommand{\cada}{\cite{sung2011}}
\newcommand{\cadb}{\cite{koppula2013}}
\newcommand{\fpadl}{\cite{pirsiavash2012}}
\newcommand{\wwadl}{\cite{bruno2014}}
\newcommand{\yg}{\cite{bullock2014}}
\newcommand{\survey}{\cite{chaquet2013}}
\newcommand{\lfd}{\cite{billiard2008}}

\newcommand{\ra}[1]{\renewcommand{\arraystretch}{#1}}
\begin{document}


\maketitle
\thispagestyle{empty}
\pagestyle{empty}

\begin{abstract}
A dataset is crucial for model learning and evaluation. Choosing the right dataset to use or making a new dataset requires the knowledge of those that are available. In this work, we provide that knowledge, by reviewing twenty datasets that were published in the recent six years and that are directly related to object manipulation. We report on modalities, activities, and annotations for each individual dataset and give our view on its use for object manipulation. We also compare the datasets and summarize them. We conclude with our suggestion on future datasets.   

\end{abstract}

\section{Introduction}

Datasets are valuable in various scientific fields because they are crucial for testing an algorithm. The demands for datasets follow the advancement of a field or the evolution of a problem, and new datasets never stopped being created. A good dataset may not only verify or deny the correctness and effectiveness of an algorithm, but may also help expose the flaws or exemplify the strength of the algorithm. To choose the good dataset, one first needs to know what datasets are available, what they include, and how they differ. Then one can decide on whether any datasets would be useful and which one or several would best serve the research purpose. One may also decide that none of the datasets suits the purpose, and the reason on which that particular decision is made can be used to improve on the existing datasets and make new ones. To help one with choosing the right dataset(s) or deciding on making new datasets, we contribute a review of datasets that we consider would be useful for research on object manipulation. All the datasets were published in the recent six years, i.e., since 2009.

As the name implies, an object manipulation motion involves an object. It intends to accomplish a certain task by manipulating, or changing the position and orientation of the object. In contrast to a gross motion such as waving and stretching, an object manipulation motion is a fine motion, and the body parts involved cover a much smaller physical space. We report on datasets that \emph{focus} on object manipulation motion. Gross motions may be present in certain datasets, but do not play the dominant role.

\begin{figure}[h]
	\includegraphics[trim={1.7cm, 15.8cm, 3.5cm, 1.8cm}, clip, width=\linewidth]{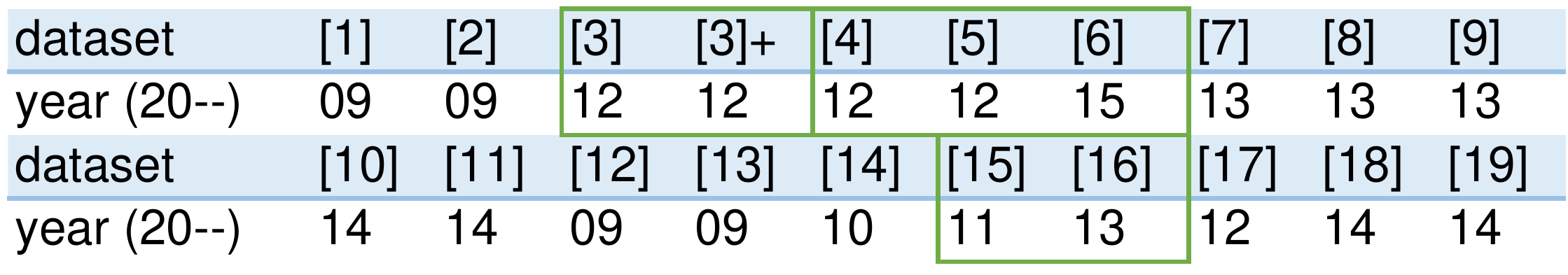}	
	\caption{The chronological order in which the datasets are presented}
	\label{fig-order}
\end{figure}

We divide the datasets into two categories and present them separately: those that include mostly cooking activities, in section \ref{sec-cooking}, and those that include more general activities of daily living (ADL), in section \ref{sec-adl}. In both categories, we sort the datasets in ascending chronological order. We keep datasets that belong to the same series together, and use the earliest publication year among the members for sorting. Fig. \ref{fig-order} shows the year of each dataset in the presented order, in which the series are delimited in green: (\gaze{}, \gazep{}), (\mpiic{}, \mpiip{}, \mpiid{}), and (\cada{}, \cadb{}). 

We downloaded each dataset and verified the consistency of the contents with the publication. When we encountered confusion or uncertainty about certain contents, we did not assume or guess but asked the authors for clarification. For each dataset, we report on the modalities, the activities performed, and annotations, then we give our view on how the dataset relates to object manipulation. After reporting on the datasets one-by-one, we summarize them on the availability of modalities, object identifiability in annotated activities, and the forms of temporal segmentation of annotated activities. We also provide the lists of shared annotated activities for the ADL and cooking datasets, respectively.

Because of the limitation in space, this review cannot be exhaustive in width or depth. To learn about datasets on more general human actions, one is directed to \survey{}. For those who wants to learn more about certain datasets covered in this work, we provide the link to each dataset in Table \ref{table-link}.


\begin{figure*}
	\begin{center}
		\includegraphics[trim={0.6cm, 6.3cm, 5.8cm, 1.9cm}, clip, width=0.9\linewidth]{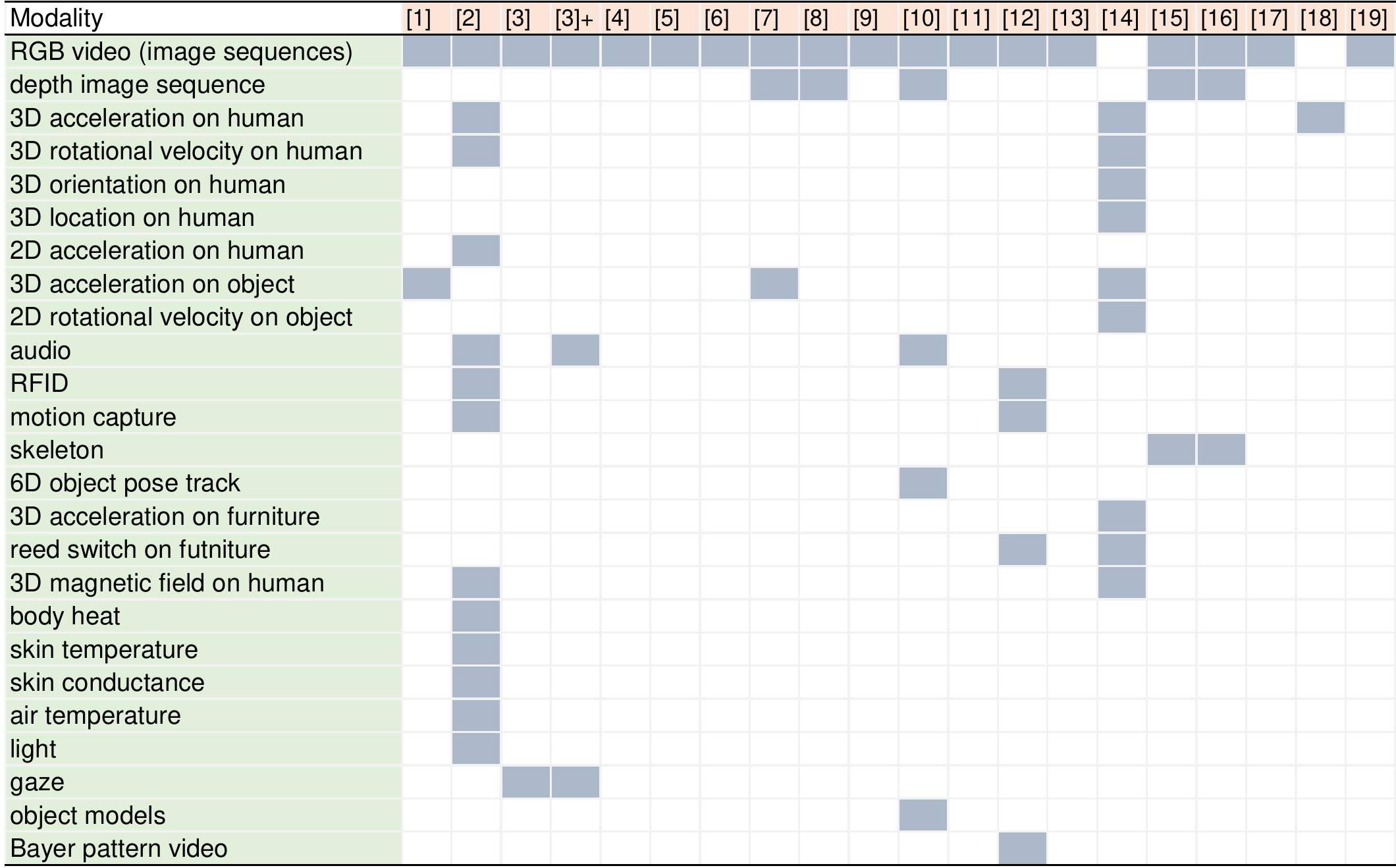}
		\caption{Modalities}
		\label{fig-modality}
	\end{center}
\end{figure*}
\begin{figure}
	\includegraphics[width=\linewidth]{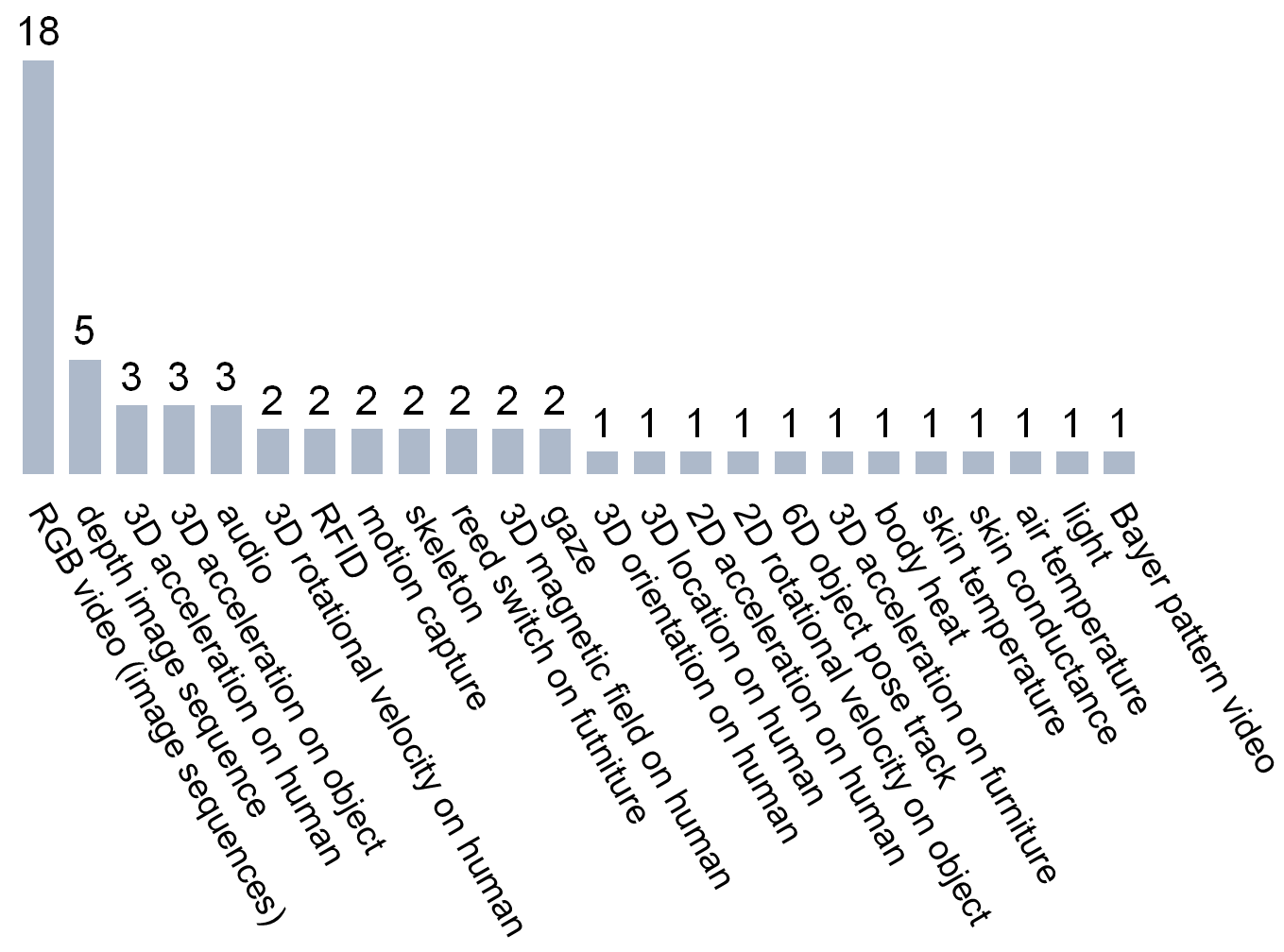}
	\caption{Count of datasets for each modality}
	\label{fig-count-modality}
\end{figure}


\section{Datasets of cooking activity} \label{sec-cooking}

\subsection{Slice\&Dice}
Slice\&Dice \sd{} features four instrumented utensils which include three knives of different sizes and a spoon. Each utensil embeds in its handle a 3-axis accelerometer. Twenty subjects each prepared a salad or a sandwich freely using the ingredients provided by the experimenter. The acceleration data are accompanied by RGB videos. We consider embedding accelerometers inside objects a merit for, unlike images, acceleration data belong to a certain object alone, and is readily usable without running object recognition first.  

\subsection{CMU-MMAC}

The CMU-MMAC dataset \mmac{} contains multi-modal cooking activities of five recipies: brownie, eggs, pizza, salad, and sandwich. The modalities include RGB videos from static and wearable cameras, multi-channel audios, motion capture, inertial measurement units (IMU), RFID, etc. We are not positive on the number of subjects that were involved, but we infer that it is between thirty-nine and forty-five. Each subject performed all the recipes. The dataset also specifically recorded anomalous accidental events that happened while cooking. Certain modalities are incomplete for certain recipes performed by certain subjects. Annotations exist for sixteen subjects making brownies and correspond to the videos captured by the wearable camera. The annotations apply the structure of ``verb$+$objectOne$+$preposition$+$objectTwo", whose components are assembled using grammar. 

Except RFID tagging which merely reports the involvement of certain objects, all modalities are on human, which is contrary to Slice\&Dice \sd{}. The dataset is rich in data of upper arm motions because of the combined use of motion capture and IMUs, and therefore is suitable for 3D manipulation motion analysis.    

\subsection{Gaze and Gaze+}

The Gaze dataset \gaze{} contains RGB egocentric videos of fourteen subjects making meals using provided ingredients on a table. The videos were captured using an eye-tracking camera and therefore are accompanied by gaze data. The Gaze+ dataset \gaze{} (later referred to as \gazep{}) is an upgrade to Gaze, and provides the two modalities in Gaze plus audio. The videos have higher resolution than Gaze, and were captured in an instrumented kitchen instead of on a simple table. Ten subjects performed seven dishes. Actions and objects were annotated in the same way as in Gaze. Compared to static images, egocentric images have much larger proportions of the image showing object manipulation specifically and contain more detail, which we consider a merit. Analyzing object motion, however, would assume that object tracking has been done. 

\subsection{The MPII Cooking dataset, Cooking Composite dataset, and Cooking 2 dataset}

MPII sequentially created three datasets related to cooking: the MPII Cooking dataset \mpiic{} which focuses on fine grained activity, the MPII Cooking Compositite dataset \mpiip{} which focuses on composite activities composed of basic-level activities, and the MPII Cooking 2 dataset \mpiid{} which unifies and is an upgrade of both \mpiic{} and \mpiip{}.   

The MPII Cooking dataset involved twelve subjects each preparing one to six out of fourteen dishes, and contains forty-four RGB high-definition (HD) videos with a total length of over eight hours or 881,755 frames. The annotations include sixty-five activities, and 5,609 instances were identified. 

The MPII Cooking Composite dataset included all the videos from the MPII Cooking dataset and added 212 newly-recorded videos. Eighteen more subjects than in the MPII Cooking dataset participated. Different from the MPII Cooking dataset, the MPII Cooking Composite dataset annotations include four categories: activities (e.g. verbs), ingredients, tools, and containers, which combined are referred to as ``attributes". There exists 218 attributes in the dataset, among which seventy-eight are activities. A total of 49,258 attribute instances have been identified which belong to 12,642 annotated temporal segments. 

As a refined superset of \mpiic{} and \mpiip{}, the MPII Cooking 2 dataset contains 273 videos involving thirty subjects. The dataset contains fifty-nine dishes, which consist of fourteen diverse and complex dishes from \mpiic{}, and forty-five shorter and simpler composite dishes from \mpiip{}. A total of 222 attributes exist, among which eighty-seven are activities. 54,774 attribute instances have been identified which belong to 14,105 temporal segments.  

For the above MPII datasets, the subjects were only told which dish to prepare, which lead to natural activities with much variability. 

Of all the datasets we include in this work, the MPII datasets altogether have the largest number of HD videos and annotation instances. Objects and fine actions are annotated in great detail, and 2D poses of upper body are also provided. For vision-based 2D object manipulation analysis, the amount of data and action variability of the MPII datasets can only be rivaled by the Brown breakfast dataset \bb{}, if not unmatched.

\subsection{50 Salad} 

The 50 Salad dataset \salad{} extends Slice\&Dice \sd{} by using accelerometers on more utensils and  by including depth videos in addition to RGB ones. Twenty-five subjects each prepared a mixed salad twice, and in each run followed a specific sequence of tasks. The sequences were produced by a statistical activity diagram, which would theoretically enable the same number of samples for each task sequence. 

The annotation includes three high-level activities: prepare dressing, cut and mix ingredients, and serve salad. Each high-level activity summarizes several low-level activities, and each low-level activity has -pre, -core, and -post phases, which were annotated respectively. 

50 Salad inherits the merit of Slice\&Dice \sd{}, involves more subjects, enables 3D analysis with depth videos, and has finer annotations. In that regard, we recommend 50 Salad over Slice\&Dice. 

\subsection{The Actions for Cooking Eggs dataset (ACE)}

The ACE dataset \ace{} contains RGB-D videos of cooking activities for five egg menus, all of which were cooked by each of seven subjects. The labels contain only verbs: break, mix, bake, turn, cut, boil, season, and peel. We include this dataset because it provides fine object manipulation motion, but since objects are not identified in any way, using the dataset would rely on human and object tracking more heavily than other datasets. 

\subsection{The YouCook dataset}

The YouCook dataset \yc{} consists of eighty-eight RGB cooking videos downloaded from Youtube. All the videos have a third person point of view.  Although only seven actions labels are used, as many as forty-eight object labels spanning seven object categories exist, and object tracks are provided. We consider the richness of object labels and the availability of the objects tracks as the merits of the dataset, of which the latter facilitates analysis of fine motion in 2D.

\subsection{The dataset of actions for making cereal} 

This dataset \kthc{} recorded eight subjects making cereal. The dataset includes multiple modalities, including  RGB-D videos, audios, estimated six degree-of-freedom (DOF) object pose trajectories, and object mesh models. We consider the object pose trajectories as the merit of the dataset. No other datasets that we include provide such modality, and using the trajectories alone suffices to conduct analysis on 3D object manipulation.

\subsection{The Brown breakfast actions dataset}

The Brown breakfast dataset \bb{} contains roughly seventy-seven hours of RGB videos involving fifty-two subjects captured at up to eighteen distinct kitchens. In total ten recipes were performed and each subject was reported to have performed all ten recipes, but available data for different subjects vary. Forty-eight coarse activity annotations exist and 11,267 annotation instances were identified.  The statistics of the dataset makes it a possible rival of the MPII datasets. It has the largest number of video frames (non HD) among the datasets we include, roughly more than the MPII datasets by 50\%. The number of coarse annotation instances is not much lower than the MPII datasets, but the detail and richness of the annotation could not compete with MPII. The dataset does include fine activity annotations, but the statistics and the description of the formation of such annotations are not yet available. Compared with MPII, the dataset lacks 2D upper body pose annotations.

\section{Datasets of activities of daily living (ADL)} \label{sec-adl}

\subsection{The TUM Kitchten dataset }

The TUM Kitchen dataset \tumk{} contains multi-modal data of set-a-table activities. The modalities include RGB and raw Bayer pattern videos, motion capture, RFID, and reed sensor. Four subjects each transported certain objects from the cupboard, the counter, and the drawer, to a table, and then laid them out in a specified way. The subjects transported objects one by one as a robot would do, and also several objects at a time as naturally done by a human. The dataset also includes repetitive activities of picking up and putting down objects. The annotations cover the entire duration of the set-a-table activity which starts with \emph{Reaching} through \emph{ReleaseGraspOfSomething}. The actions of the left hand, the right hand, and the trunk were annotated respectively. 

Similarly to CMU-MMAC \mmac{}, the dataset identifies objects involved during motion execution, and the availability of motion capture makes it a good candidate for 3D analysis on pick-and-place motion.

\subsection{The Rochester ADL dataset}

The Rochester ADL dataset \radl{} contains RGB videos of five subjects performing certain ADL and Instrumented ADL (IADL) activities which can be summarized as: using phone, writing, drinking and eating, and preparing food. Each video records one activity. Similar to the MPII datasets \mpiic{}-\mpiid{} and Brown breakfast dataset \bb{}, the Rochester ADL dataset would rely on human and object recognition to be useful for 2D fine motion analysis.

\subsection{The OPPORTUNITY dataset}

The OPPORTUNITY dataset \opp{} contains multi-modal data of five morning ADL runs and one Drill run for each of four subjects. Motion sensors were densely deployed on human body, on the objects, and in the environment. The modalities on human body include IMUs, 3D accelerometers, and 3D localizers. The modalities on objects include 3D accelerometers and 2D rotational velocity sensors. The annotations consists of five ``tracks": locomotion, high-level activity, mid-level gestures, low-level actions and objects for the left and the right hand, respectively. 

The dataset distinguishes itself from others that we include by using accelerometers and rotational velocity sensors on \emph{both} hand and objects. Since object manipulation analysis focuses on the interaction between hand and objects, data that include the motion of both hand and objects are desired. The dataset is comparable with 50 Salad \salad{}, CMU-MMAC \mmac{}, and TUM Kitchen \tumk{} in modality availability, although the last three target cooking scenarios. For the objects, the dataset includes 2D rotational velocity, which is unavailable in 50 Salad. For the human body, the dataset lacks motion capture, which is available in CMU-MMAC and TUM Kitchen, but alternatively provides 3D acceleration and 3D rotational velocity.

\subsection{The Cornell CAD-60 and CAD-120 datasets}

The CAD-60 \cada{} and the CAD-120 \cadb{} are both RGB-D video datasets. CAD-60 includes video sequences of four subjects performing twelve ADLs in five different indoor environments. Each sequence corresponds to one instance of a certain activity. The CAD-120 dataset recorded four subjects each performing ten high-level activities. Each subject performed every high-level activity multiple times with different objects. The annotations include ten low-level activities, and twelve object affordances.

CAD-60 and CAD-120 feature skeleton data, which include tracks of 3D position of all fifteen joints plus 3D orientation of eleven joints. The skeleton is similar to motion capture, but with much fewer defined joints and less corresponding data. Despite the ``lightness" compared with motion capture, the skeleton is directly usable for 3D fine motion analysis, and therefore we consider it as the merit of the datasets.  

\subsection{The first person ADL dataset}

The ADL dataset by Pirsiavash \fpadl{} contains RGB videos captured using a GoPro camera. It recorded twenty subjects performing eighteen ADLs. Forty-two objects were annotated by annotators with bounding boxes, tracks, and the status as to whether the object is being interacted with. Similar to Gaze(+) \gaze{}, with first person images, the working area of the hands is emphasized. However, since the dataset includes a single modality, using it for analysis on 2D fine motion would rely on object tracking.

\subsection{The wrist-worn accelerometer dataset} 

The wrist-worn accelerometer dataset \wwadl{} contains accelerometer data of sixteen subjects performing a total of fourteen ADLs. The accelerometers were attached to the right wrists of the subjects and the data were recorded at the subjects' home. The dataset contains 979 trials. For fine motion analysis, wrist acceleration may be less ideal than hand acceleration, but it remains a readily usable modality.

\subsection{The Yale human grasping dataset}

The Yale human grasping dataset \yg{} contains 27.7 hours of RGB wide-angle videos of profession-related manipulation motion. Two machinists and two housekeepers participated. The dataset is intended for grasping analysis. The annotations were done on two levels. On the first level, the grasp type was annotated along with the corresponding task name and object name. The second level provided the properties of the object and the task. A total of 18,210 grasp instances have been annotated. The dataset includes prolonged videos of manipulation motion of machining and housekeeping alone, two categories that are not to be found in other datasets that we include.            

\section{Dataset Summary} \label{sec-summary}

\newcommand{\has}{$\bullet$}
\newcommand{\colwidtt}{p{0.1\textwidth}}
\begin{table*}
	\begin{threeparttable}
		\ra{1.5}
		\caption{Shared annotated ADLs. }
		\label{table-adl}
		\begin{tabular}{@{} p{2.5cm} \colwidtt \colwidtt \colwidtt \colwidtt \colwidtt \colwidtt \colwidtt @{}}
			\hline
			Activities              &  \tumk     & \radl      & \opp        & \cada     & \cadb       & \fpadl       & \wwadl        \\
			\hline
			use phone               &            & answer phone, dial on a phone      
			&             & talk on the phone      
			&             & \has         & \has        \\
			
			write on whiteboard     &            & \has       &             &  \has     &             &              &      \\
			
			drink                   &            &   \has     & sip         & \has      & \has        & \has         & \has   \\
			
			eat                     &            & \has       &             &           & \has        &              & \has   \\
			
			chop/cut                &            & chop       & cut         & chop      &             &              &     \\
			
			reach                   &  \has      &            & \has        &           & \has        &              &     \\
			
			release             &  release grasp &            & \has        &           &             &              &     \\
			
			comb hair               &            &            &             &           &             & \has         & \has         \\
			
			brush teeth             &            &            &             &  \has     &             & \has         &  \has       \\
			
			use computer            &            &            &             &  \has     &             &  \has        &         \\
			
			move                    &            &            &             &           &  \has       &  dishes      &         \\
			
			stir                    &            &            & \has        &  \has     &             &              &         \\
			
			pour                    &            &            &             &           &   \has      &              &  \has        \\
			
			open                 & door, drawer  &            & \has        &           &   \has      &              &         \\
			
			close                & door, drawer  &            & \has        &           &   \has      &              &         \\
			\hline

		\end{tabular}
		\begin{tablenotes}
			\item We only consider low-level annotations for \opp{}.

		\end{tablenotes}
	\end{threeparttable}
\end{table*}

Fig. \ref{fig-modality} lays out the different modalities included in each dataset, and Fig. \ref{fig-count-modality} shows in descending order the count of datasets for each modality. We can see from the figures that, as the most easily managed modality, RGB video leads with eighteen datasets excluding only \opp{} and \wwadl{}. In fact, \opp{} did collect RGB videos but did not publish them. Depth video is the second most adopted modality, but with a count much lower than that of RGB video. 3D acceleration (on human or object) leads among the rest modalities, but no modalities besides videos stand out. Motion capture data are only found in \mmac{} and \tumk{}, possibly because of the cost and effort required in the setup of the system (although \tumk{} uses a markerless capture system and most of the effort is with computing). The skeleton tracks, which can be considered as a light version of the motion capture, are only available in \cada{} and \cadb{}.  

Research in object manipulation might find 3D object poses very useful. Acceleration and rotational velocity may be used to estimate object poses, but explicit or readily usable recordings of object poses, which may require a motion capture system, are unavailable in the datasets. \kthc{} is the only dataset that provides something close: the \emph{estimated} 6 DOF object pose trajectories. Object motions that are simpler than poses can be obtained if a sensor actively takes samples and is attached to an object. Datasets with such setup include
\begin{enumerate}
	\item \sd{}, \salad{}. Objects were equipped with accelerometers. 
	\item \opp{}. Objects were equipped with accelerometers and rotational velocity sensors. Furniture and appliances were equipped with reed switches and accelerometers. 
	\item \tumk{}. Doors were equipped with reed switches. 
\end{enumerate}

Activity annotations can be useful for various purposes. We identified the annotated activities that are shared by multiple datasets, and list those belonging to the ADL datasets in Table \ref{table-adl}, and those belonging to the cooking datasets in Table \ref{table-cooking}.  In both tables, we combine similar annotations and specify each in the cells. For example, on the first row of Table \ref{table-adl}, the annotated activity is summarized as ``use phone", whereas \radl{} specifically uses ``answer phone" and ``dial on a phone", and \cada{} specifically uses ``talk on the phone". 

The shared activities show the consensus among different authors on what activities should be performed and annotated, which can be helpful for one who tries to make such decision when making a new dataset. However, because the amount of authors is limited, not being a shared activity does not necessarily mean the activity is not important. Therefore, we also provide the complete list of annotated activities at \url{http://rpal.cse.usf.edu/motiondatasetreview/index.htm}, for cooking and ADL, respectively. The shared activities can also help with using more than one dataset. If one wants to study a certain shared activity, one could use several datasets that include this activity together to access more modalities and higher variability.    
	




Except for annotated activities, objects that are involved in an activity may also  be helpful for activity analysis. For all datasets except \ace{}, objects are identifiable in the annotated activities through
\begin{enumerate}
	\item being separately annotated: \mpiip{}, \mpiid{}, \yc{}, \cadb{}, \fpadl{}, \yg{}. 
	\item being part of the annotation phrases: \sd{}, \mmac{}, \gaze{}, \gazep{}, \mpiic{}, \salad{}, \kthc{}, \bb{}, \tumk{}, \radl{},  \cada{}, \wwadl{}. 
	\item being equipped with sensors
	\begin{enumerate}
		\item accelerometers: \sd{}, \salad{}, \opp{}.
		\item rotational velocity sensors: \opp{}.
		\item reed switches: \tumk{}, \opp{}.
		\item RFID: \mmac{}, \tumk{}. 
	\end{enumerate} 
\end{enumerate}

Temporal segmentation of annotated activities is also important for activity analysis. For \yg{}, temporal segmentation does not apply because \yg{} focuses on grasp instances. All other datasets include temporal segmentation, in the following forms
\begin{enumerate}
	\item video subtitle: \sd{}, \kthc{}.
	\item explicit video time: \gazep{}, \fpadl{}. 
	\item frame number: \mmac{}, \gaze{}, \mpiic{}, \mpiip{}, \mpiid{}, \ace{}, \yc{}, \bb{}, \tumk{}, \cadb{}.  
	\item timestamp: \salad{}, \opp{} 
	\item implicit: \radl{}, \cada{}, \wwadl{}.
\end{enumerate}

We are aware of the existence of other related datasets, however, to keep this work focused we do not include them. Examples of the excluded datasets are
\begin{enumerate}
	\item \survey{}, and \cite{soomro2012}, \cite{huynh2008}, which are datasets that do not include object manipulation motions, or if they do, the object manipulation motions are sparse. 
	\item \cite{lai2011}, \cite{singh2014}, and \cite{calli2015}, which are dataset of objects that are typically involved in manipulation, rather than datasets of motion.
\end{enumerate}

Most datasets we include are intended for action recognition. However, researchers who work on learning from demonstration (LfD) \lfd{} intend to reproduce human actions rather than recognizing them. Thus, we suggest that except for choosing from the modalities we have reviewed, a more ideal dataset for LfD should also aim to provide readily usable data that are more closely related to dynamic and kinematic motion execution. Examples of suggested modalities include trajectories of object poses, joint poses of human upper body, hand posture, torque, force between hand and object, etc.  

\section{Conclusion}

We reviewed twenty datasets that we considered useful for research on object manipulation. We reported on each dataset individually, gave our view on the relation between each dataset and object manipulation, and compared and summarized all of them together. We provided suggestion on future datasets, and we are putting that suggestion into practice and making a new dataset.   

\section{Acknowledgement}
This material is based upon work supported by the National Science Foundation under Grant No. 1421418.

\begin{table}
	\begin{threeparttable}
		\caption{Link to datasets}
		\label{table-link}
		\begin{tabular}{@{} p{0.04\textwidth} p{0.42\textwidth} @{}}
			\hline
			\sd     & \url{http://openlab.ncl.ac.uk/publicweb/publicweb/AmbientKitchen/KitchenData/Slice&Dice_dataset/}\\
			\mmac   &   \url{http://kitchen.cs.cmu.edu/}\\
			\gaze(+) &  \url{http://ai.stanford.edu/~alireza/GTEA_Gaze_Website/}\\
			\mpiic  &   \url{https://www.mpi-inf.mpg.de/departments/computer-vision-and-multimodal-computing/research/human-activity-recognition/mpii-cooking-activities-dataset/}\\
			\mpiip  &   \url{https://www.mpi-inf.mpg.de/departments/computer-vision-and-multimodal-computing/research/human-activity-recognition/mpii-cooking-composite-activities/}\\
			\mpiid  &  \url{https://www.mpi-inf.mpg.de/departments/computer-vision-and-multimodal-computing/research/human-activity-recognition/mpii-cooking-2-dataset/}\\
			\salad  & \url{http://cvip.computing.dundee.ac.uk/datasets/foodpreparation/50salads/}\\
			\ace    &   \url{http://www.murase.m.is.nagoya-u.ac.jp/KSCGR/}\\
			\yc     &   \url{http://www.cse.buffalo.edu/~jcorso/r/youcook/}\\
			\kthc   &   \url{http://robocoffee.org/datasets/}\\
			
			\bb     & \url{http://serre-lab.clps.brown.edu/resource/breakfast-actions-dataset/}\\
			\tumk   &   \url{https://ias.in.tum.de/software/kitchen-activity-data}\\
			\radl   &   \url{http://www.cs.rochester.edu/~rmessing/uradl/}\\
			\opp    &   UCI repository: \url{https://archive.ics.uci.edu/ml/datasets/OPPORTUNITY+Activity+Recognition#}, Challange: \url{http://www.opportunity-project.eu/challengeDataset} \\
			\cada{}\cadb &  \url{http://pr.cs.cornell.edu/humanactivities/data.php}\\ 
			\fpadl  &   \url{http://people.csail.mit.edu/hpirsiav/codes/ADLdataset/adl.html}\\		
			\wwadl  &   \url{https://archive.ics.uci.edu/ml/datasets/Dataset+for+ADL+Recognition+with+Wrist-worn+Accelerometer#} \\
			\yg     &   \url{http://www.eng.yale.edu/grablab/humangrasping/}\\
			
			\hline			
		\end{tabular}
		\begin{tablenotes}
			\item \gaze{}(+) refers to both Gaze and Gaze+ 
		\end{tablenotes}
	\end{threeparttable}	
\end{table}

\bibliographystyle{IEEEtran}
\bibliography{motion_list}

\newcommand{\colwidthb}{p{0.09\textwidth}}

\onecolumn
\begin{landscape}
\begin{table*}
	\begin{threeparttable}
	\ra{1.8}
	\caption{Shared annotated cooking activities}
	\label{table-cooking}
	\begin{tabular}{@{} p{3.5cm} \colwidthb \colwidthb \colwidthb \colwidthb \colwidthb \colwidthb \colwidthb \colwidthb \colwidthb \colwidthb @{}}
		\hline
Activity
&  \sd          & \mmac      & \gaze        & \gazep      & \mpiid        & \salad      & \ace        &\yc         & \kthc        &\bb   \\
\hline
chop/cut        
& chop, slice, dice &        &              & cut         & chop, cut, cut apart, cut dice, cut off ends, cut off inside, cut stripes, slice
                                                                          & cut         & cut         &            &             & cut    \\
                                                                            
peel/shave
& peel, shave  &             &              & peel        & peel          & peel        & peel        &            &             &  peel    \\  

stir/mix
& stir         & stir        &              & mix         & mix, stir     &  mix        & mix         & stir       &             &  stir    \\ 

pour
&              &  \has       & \has         & \has        & \has          &             &             & \has       & milk, cereal & \has    \\
 
put/place
&              & put         &              & put         & put in, put on & place      &             & put down   &              & put    \\
 
take
&              & \has        & \has         & \has        & take lid, take out &        &             &            &              & \has   \\
                                                                           
spread/smear
& spread       &             & spread       & spread      & spread        &             &             &            &              & smear   \\

eat/taste
& eat          &             &              &             & taste         &             &             &            &              &      \\

scoop/spoon
& scoop        &             & scoop        &             &               &             &             &            &              & spoon     \\

season/spice
&              &             &              &             &  spice        &             & season      & season     &              &         \\

turn/flip
&              &             &              &  flip       &  turn over    &             & turn        & flip       &              &         \\

open/close food (container)
&              &  open       &  \has        &  \has       &  \has         &             &             &            & \has         &         \\

open/close drawer
&              &  open       &              &             &  \has         &             &             &            &              &         \\

open/close dishwasher/oven
&              &             &              & oven        &  \has         &             &             &            &              &         \\

open/close cupboard /fridge /microwave
&              &  \has       &              & fridge      & \has          &             &             &            &              &         \\

crack/break
&              & egg         &              & \has        & open egg      &             &  \has       &            &              & \has        \\

beat/whip
&              & beat egg    &              &             & whip          &             &             &            &              &         \\         

add
&              &             &              &             &  \has         & \has        &             &            &              &  teabag, salt and pepper, topping \\ 

squeeze
&              &             &              & \has        & \has          &             &             &            &              & \has  \\

turn on/off
&              &             &              & \has        & \has          &             &             &            &              &  \\

wash
&              &             &              & \has        & \has          &             &             &            &              &  \\

dry
&              &             &              & \has        & \has          &             &             &            &              &  \\

fill
&              &             &              & \has        & \has          &             &             &            &              &  \\

\hline

	\end{tabular}
	\begin{tablenotes}
		\item Since \mpiid{} supercedes \mpiic{} and \mpiip{}, we only include \mpiid{} in the table. 
	\end{tablenotes}
	\end{threeparttable}
\end{table*}
\end{landscape}

\twocolumn

\end{document}